\documentclass[twocolumn]{svjour3}          
\smartqed  
\usepackage[usenames]{color}
\usepackage[numbers]{natbib}
\usepackage[table]{xcolor}
\usepackage{amssymb}
\usepackage{makecell}
\usepackage{multicol}
\usepackage{multirow}
\usepackage{enumitem}
\usepackage{rotating}
\usepackage{array}
\usepackage{times}
\usepackage{graphicx}
 \usepackage{psfrag}
\usepackage{subfigure}
\usepackage{rotating}
\usepackage{url}
\usepackage{rotating, graphicx}
\usepackage{amsmath}
\usepackage{booktabs}
\usepackage{lscape}
\usepackage{verbatim}
\usepackage{geometry}
\usepackage[misc]{ifsym}
\papertype{generic article}
\usepackage{breakurl}
\usepackage{graphicx}
\usepackage[pagebackref=true,breaklinks=true,linkcolor=red,anchorcolor=blue, citecolor=green,letterpaper=true,colorlinks,bookmarks=false]{hyperref}
\usepackage{pifont}
\usepackage[font={small}]{caption}
\usepackage{mathrsfs}
\usepackage{svg}
\usepackage{calc}
\usepackage{wrapfig}
\usepackage[ruled,linesnumbered]{algorithm2e}

\definecolor{DarkBlue}{rgb}{0.0,0.08,0.6}
\definecolor{DarkRed}{rgb}{0.6,0.00,0.08}
\definecolor{DarkGreen}{rgb}{0.0,0.6,0.08}
\definecolor{LightBlue}{rgb}{0.88,0.92,0.95}
\definecolor{Orange}{rgb}{1,0.75,0}

\begin{document}

\title{Sample-aware RandAugment: Search-free Automatic Data Augmentation for Effective Image Recognition}


\author{Anqi Xiao $^{1}$  \and
        Weichen Yu $^{2}$ \and
        Hongyuan Yu $^{3}$ $\textrm{\Letter}$ 
}

\authorrunning{Anqi Xiao \emph{et al.}} 

\institute{ Anqi Xiao (xaq364856589@gmail.com) \\
        Weichen Yu (wyu3@andrew.cmu.edu) \\
        $\textrm{\Letter}$ Hongyuan Yu (yuhyuan1995@gmail.com) \\
       1 University of Chinese Academy of Sciences, China \\
       2 Carnegie Mellon University, USA \\
       3 Xiaomi Inc., China \\
 }

 \date{Accepted by \textit{International Journal of Computer Vision}, 22 July 2025 }

\maketitle

\begin{abstract}
Automatic data augmentation (AutoDA) plays an important role in enhancing the generalization of neural networks. However, mainstream AutoDA methods often encounter two challenges: either the search process is excessively time-consuming, hindering practical application, or the performance is suboptimal due to insufficient policy adaptation during training. To address these issues, we propose Sample-aware RandAugment (SRA), an asymmetric, search-free AutoDA method that dynamically adjusts augmentation policies while maintaining straightforward implementation. SRA incorporates a heuristic scoring module that evaluates the complexity of the original training data, enabling the application of tailored augmentations for each sample. Additionally, an asymmetric augmentation strategy is employed to maximize the potential of this scoring module. In multiple experimental settings, SRA narrows the performance gap between search-based and search-free AutoDA methods, achieving a state-of-the-art Top-1 accuracy of 78.31\% on ImageNet with ResNet-50. Notably, SRA demonstrates good compatibility with existing augmentation pipelines and solid generalization across new tasks, without requiring hyperparameter tuning. The pretrained models leveraging SRA also enhance recognition in downstream object detection tasks. SRA represents a promising step towards simpler, more effective, and practical AutoDA designs applicable to a variety of future tasks. Our code is available at \href{https://github.com/ainieli/Sample-awareRandAugment}{https://github.com/ainieli/Sample-awareRandAugment}.


\end{abstract}

\section{Introduction}\label{sec1}

Automatic data augmentation (AutoDA) is ubiquitous in training methods, capable of autonomously adjusting and exploring optimal augmentation policies tailored to various target tasks \citep{AA}. 
It enhances the generalization of neural networks for image recognition by filling in the missing data in the target distribution \citep{FastAA,DA_fill_miss_data}.
However, current AutoDA methods typically suffer from excessive search time required prior to application in training \citep{AA,AWS,RA}, or the implementation of dynamically adjusting policies that come at the expense of significant search overheads \citep{MetaA,AdvAA,LatentA}. Additionally, complex optimization strategies present obstacles that hinder widespread adoption and limit the applicability of AutoDA across different tasks \citep{DADA,DDAS,Invariance_AutoDA}. 

An emerging trend in the field of AutoDA is the pursuit of methods that balance simplicity with effectiveness. With the advent of policy transferring strategies \citep{AA,FastAA}, AutoDA can be applied without notable performance degradation. In addition, substantial reductions in search space also allow manual tuning to avoid time-consuming search from scratch \citep{RA}. These advancements boost the development of search-free AutoDA methods \citep{TA,UA,timm}. Search-free augmentation strategies show great potential by generating randomly augmented samples of original images. Nevertheless, their ability to achieve peak performance is often constrained by their inherent simplicity \citep{RA,TA}. 

We identify two primary issues prevalent in current AutoDA methods: 1) For search-based methods, the complicated time-consuming search process sets barriers to broader applications, and 2) For search-free methods, suboptimal performance often results from the deficient flexibility to adapt and adjust the policy dynamically during training. We find that search-free AutoDA methods are favored in practical scenarios (such as the training of new types of architectures) due to their simplicity for adoption, which motivates us to design a flexible, search-free method capable of generating effective input for image recognition without substantially increasing the complexity and cost of the training process. To this end, we develop a search-free, sample-aware AutoDA method named Sample-aware RandAugment (SRA). We hypothesize that the core idea of improving the performance of search-free AutoDA hinges on focusing on samples that are particularly valuable for determining the decision boundaries during training \citep{importance_of_learn_boundary, importance_of_learn_boundary_2}. To achieve this, we propose an asymmetric training strategy that divides the original batch into two sub-batches, applying distinct augmentation policies for exploration and refinement, respectively. During the exploration phase, random augmentation is used to explore the target data distribution and improve the representation ability of the target model. While in the refinement phase, a heuristic module Magnitude Instructor Score (MIS) is proposed to assess the difficulty of each sample in the sub-batch, which enables SRA to generate more hard samples that are crucial for refining the decision boundaries. Both sub-batches are used to update target model weights.

The proposed SRA is evaluated on CIFAR \citep{CIFAR} and ImageNet \citep{ImageNet} benchmarks. Experiments on convolutional neural networks (CNN), vision Transformers, and newly rising state space model-based vision Mamba architectures demonstrate that SRA outperforms existing search-free AutoDA methods in diverse settings. Moreover, it achieves competitive or even better performance compared to search-based state-of-the-art methods. Notably, SRA achieves a Top-1 accuracy of 78.31\% on ImageNet using ResNet-50, surpassing the search-free state-of-the-art by 0.24\% without relying on extensive tricks. Additionally, it is compatible with frameworks such as repeated augmentation \citep{BA} and multi-view contrastive learning \citep{TiedA}. Without requiring any tuning of training hyperparameters, SRA boosts accuracy on both the fine-grained image classification benchmark \citep{Food101} and the long-tailed image classification benchmark \citep{ImageNet-LT}. These results highlight that SRA has strong compatibility with existing training pipelines and robust generalization to new tasks. The pretrained model using SRA also performs well in downstream object detection tasks on the COCO dataset \citep{COCO}. As a search-free method, SRA is ready for out-of-the-box applications. The key contributions of this work are summarized as follows:

\begin{figure}[!t]
\small
\centering
\includegraphics[width=0.9\linewidth]{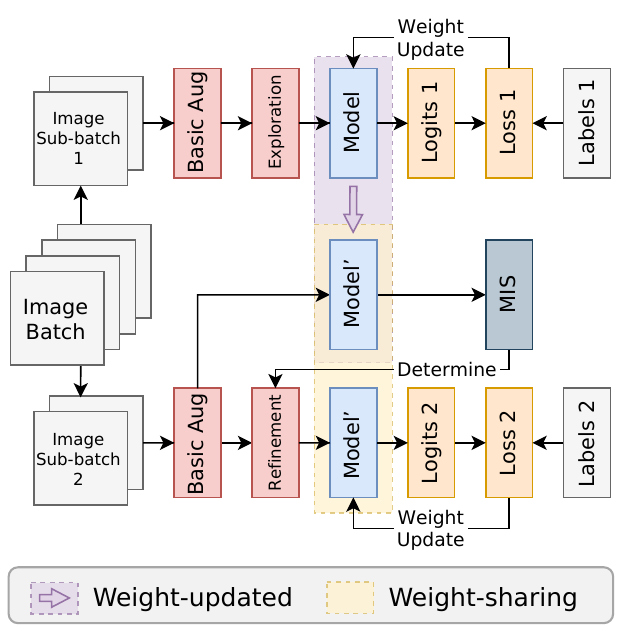}
\captionof{figure}{Training pipeline of the proposed Sample-aware RandAugment (SRA). MIS is the proposed scoring module to evaluate the difficulty of samples.}
\label{Fig: pipeline}
\end{figure}

\begin{itemize}
    \item We propose Sample-aware RandAugment (SRA), a search-free, sample-aware automatic data augmentation method that bridges the gap between well-performed yet time-consuming search-based methods and simpler, yet sub-optimal, search-free ones. SRA shows competitive or better performance compared to search-based designs in various settings, showcasing the effectiveness of the proposed search-free design.
    \item We propose a heuristic module, Magnitude Instructor Score (MIS), which dynamically evaluates the difficulty of the original training data to guide SRA in generating hard samples contributing to improved decision boundaries during training. The proposed MIS also provides new insights into sample-awareness in data augmentation, emphasizing how different samples can be prioritized for augmentation.
    \item We present an asymmetric data augmentation strategy that applies two distinct augmentation policies within a single batch, with one focusing on exploring the training data distribution and the other on refining it. This asymmetric approach introduces a novel perspective, suggesting that hybrid data augmentation designs can unlock the full potential of data augmentation in neural network training.
\end{itemize}

\section{Related Work}\label{sec2}

Automatic data augmentation (AutoDA) has emerged as a powerful technique, significantly improving performance in image recognition. It is typically controlled by a set of policy parameters that define the types of deformations, their ranges, and sampling probabilities. Compared with human-designed, widely applied augmentation methods \citep{rand_erase,Cutout,Mixup,CutMix}, AutoDA usually generate images with less synthetic semantics. The goal of AutoDA is to automatically discover an optimal augmentation strategy that fills in the missing points in the target data distribution \citep{FastAA,DA_fill_miss_data}, thereby enhancing the generalization of neural networks. 

The development of AutoDA arises from extremely time-consuming search-based methods that require hundreds or thousands of GPU hours to find the optimized policy for the target task, even on a proxy that uses a subset or smaller model for policy estimation \citep{AA}. The dramatic search cost is unrealistic for out-of-the-box use. Subsequent works try to improve the performance meanwhile reducing the search cost, with techniques such as Bayes optimization \citep{PBA,FastAA}, weight-sharing strategies \citep{AWS}, differentiable learning \citep{PAMI_diff_aug_review,DADA,FasterAA,DDAS,SLACK}, multi-armed bandit algorithm \citep{BDA}, out-of-distribution detection \citep{DualAug}, or simply expanding the potential augmented image space \citep{RangeA,DeepAA,FreeA}. Some AutoDA methods try to dynamically adjust the policy during training, while requiring repeatedly augmenting the same batch that obviously prolongs training time \citep{OHLAA,AdvAA,LatentA}, or requiring extra policy network \citep{AdaAugment,MADAug} or teacher model \citep{TeachAug} to choose the optimized policy. In general, search-based AutoDA methods are difficult to simultaneously achieve simplicity, cost-effectiveness, and performance advantages. 

A rising trend for designing AutoDA is simple yet effective. RandAugment \citep{RA} (RA) was intended to pursue effectiveness while keeping simplicity on the target task to avoid the bias from proxy search. Due to the dramatic reduction of the search space, RA is also manually tunable without any search. Meanwhile, TrivialAugment \citep{TA} and UniformAugment \citep{UA} generate augmentation sub-policy through random sampling, which also yields plenty of variants of the original data. Wightman et al. \citep{timm} hypothesizes magnitudes following a normal distribution in RA, therefore increasing the flexibility of the original RA. EntAugment \citep{EntAug} proposes a magnitude calculation method using information entropy to dynamically control the deformation level for each sample. These search-free methods only require tuning a few hyperparameters, which is easy to achieve through human priors. These simple heuristic designs show amazing performance that is competitive with many search-based ones, making AutoDA more suitable for wide applications. However, the heuristic designs are usually suboptimal for the target task. Therefore, search-free methods can hardly achieve the state-of-the-art performance.

Another general problem of the previously mentioned methods is target-aware, neglecting the variations within individual samples in the target task. The idea of customized augmentation boosts the development of sample-aware \citep{AdaTrans, MetaA, SelectA, InstaAug} or label-aware \citep{LA3, AdaAug} AutoDA, with which the performance further improves. In particular, AdaAug \citep{AdaAug} decouples the DA process as an adversarial learning problem and learns lable-aware augmentation policies. MetaAugment \citep{MetaA} uses a policy network for reweighting loss weights to achieve sample awareness. SelectAugment \citep{SelectA} uses two actor-critic structures to determine the suitable samples to apply augmentation. InstaAug \citep{InstaAug} uses a parametric plug-in invariance module to learn the local invariance of the input and adapts the augmentation to achieve diversity. MADAug \citep{MADAug} uses bi-level optimization with a validation set to measure current model and ensure sample-aware augmentation. These sample-aware methods mainly focus on the evaluation of importance of different samples rather than the deformation for samples. Meanwhile, they also require complicated optimization strategies to achieve policy learning, which sets barriers to easy implementation. Although EntAugment \citep{EntAug} focuses on magnitude selection for augmentation, the information entropy feedback is not utilized in real-time, which brings bias to performance.

\section{Sample-aware RandAugment}\label{sec3}

\subsection{Motivation}
Current mainstream AutoDA methods suffer from either complicated time-consuming search processes, or limited performance due to a deficiency of awareness in dynamically adapting and adjusting the policy during training. To address the two problems, we propose SRA that can simultaneously achieve search-free and sample-aware properties. We add a heuristic sample perception module Magnitude Instructor Score (MIS) that uses cosine similarity-based formula to dynamically evaluate the difficulty of the original data during training, which avoids complex search process. Meanwhile, since focusing only on hard samples may result in a biased represented distribution of the original training data \citep{DualAug}, we also introduce an asymmetric augmentation strategy that alternatively explores and refines the training data distribution to avoid severe over-fitting on hard samples. The modified training pipeline is shown in Fig. \ref{Fig: pipeline}. 
To balance the iterations and the number of processed samples of the two policies, we adopt a large batch split strategy in practice, where the batch size is twice that of the traditional one, while the number of iterations for updating model weights remains almost the same.

\subsection{Search Space}
SRA shares several similar designs with RA. To be competitive with many previous works, SRA adopts the search space from RA that contains 14 candidate augmentation operators. SRA also includes multiple augmentation operators in one sub-policy to augment a single image. The main difference in the search space between SRA and RA is that SRA uses continuous floating-point numbers for magnitudes, rather than discrete deformation levels. Similar search space designs utilizing continuous floating-point numbers have been widely adopted in previous works \citep{DADA,SLACK,FreeA,AdaAugment}, as they enhance the diversity of augmented samples. The valid ranges and names of these operators are shown in Table \ref{search_space}. Although the candidate operators can be specifically selected or sampled using learnable weights, for simplicity, we assign them equal probability of being sampled and applied.

\begin{table}[!h]
\centering
\small
\renewcommand\arraystretch{1.1}
\captionof{table}{The candidate 14 operations and corresponding valid ranges in the search space of SRA. *: Implemented using PyTorch, which is identical to the value range [4,8] when using Pillow\protect\footnotemark{}.}
\resizebox{\linewidth}{!}{
\begin{tabular}{lc|lc}
\hline
Operator & Valid Range & Operator & Valid Range                       \\ 
\hline
ShearX        & [-0.3, 0.3]                   & Sharpness     & [0.1, 1.9] \\
ShearY        & [-0.3, 0.3]                   & Contrast      & [0.1, 1.9]\\
TranslateX    & [-0.45, 0.45]                 & Solarize      & [0, 256]\\
TranslateY    & [-0.45, 0.45]                 & Posterize     & [0, 4]* \\
Rotate        & [$-30^{\circ}$, $30^{\circ}$] & Equalize      & - \\
Brightness    & [0.1, 1.9]                    & AutoContrast  & - \\
Color         & [0.1, 1.9]                    & Identity      & - \\
\hline
\end{tabular}
}

\label{search_space}
\end{table}
\footnotetext{https://github.com/python-pillow/Pillow}

\subsection{Sample-aware RandAugment}

\textbf{Asymmetric augmentation strategy.} The training pipeline of SRA is different from most previous works that augment each batch with the same strategy. Instead, an asymmetric strategy is adopted to improve generalization. In detail, SRA achieves data distribution exploration and refinement through three steps. For each iteration during training, a large batch containing two small sub-batches with a balanced number of samples is randomly sampled from the training dataset, with the first sub-batch undergoing Step 1 for exploration, and the other undergoing Steps 2 and 3 for refinement. Note that the loss is calculated and the gradients are propagated backward to update model weights for each sub-batch. Therefore, the model weights of Step 2 and 3 are shared while different from Step 1. The process is similar to meta-learning that uses nested update steps to estimate the optimal model weights and meta parameters, while different because the "meta parameters" here are model weights as well. In addition, the training data are not separated into different parts for updating model weights and meta parameters, respectively. The pseudo-code of SRA is in Algorithm \ref{SRA}.

\textbf{Step 1: Distribution exploration.} Data augmentation is expected to fill in the missing points of the training data distribution \citep{FastAA,DA_fill_miss_data}. The augmented image space is usually larger than the original one due to the complex transformations that generate variants of the samples. This expanded space is expected to cover more samples in the target distribution. Therefore, exploring the target distribution with augmented samples is important to avoid models over-fit on the original training data. Since we assume that no prior knowledge on how to generate beneficial augmented data in the target distribution is available, we adopt a random exploration strategy to generate variants of the original samples. We sample random operators from the candidate operator set and random magnitudes from a uniform distribution $\mathcal{U}(0,1)$ to transform the training data. Different augmentation operators are sequentially applied to form the sub-policy that widens the augmented image space, where $D$ is the number of augmentation operators in one sub-policy, or augmentation depth in the following. The magnitude is sampled independently for each operator within the sub-policy. Note that in RA, $D$ is obtain via grid search. On the contrary, $D$ of SRA is settled based on the empirical value of some previous AutoDA \citep{AA, FastAA} in the classical benchmarks. The impact of this parameter is shown in Table \ref{aug_depth} in the following section. Simplifying the augmentation process makes SRA focus only on the dynamic adjustment of magnitudes, which allows efficient and simplified augmentation to improve the recognition ability of neural networks.

\textbf{Step 2: Sample Perception.} We propose an intuitive sample-aware augmentation strategy where easy samples require heavier deformations, while hard samples require lighter ones, which is expected to generate plenty of samples for determining decision boundaries. Evaluating the difficulty of the original training data is necessary to control the deformation of the augmented images. A heuristic sample-aware perception module to evaluate the difficulty of the original images is proposed, which we call Magnitude Instructor Score (MIS). For convenience, this score is directly applied as the magnitude for the augmentation operators. To be directly applied in data augmentation, the score requires two features: 1) The range of the value should be within value range $[0, 1]$; and 2) Easier samples should have larger scores, while harder samples smaller. 

To satisfy the two demands, we simply choose cosine similarity as the basis of MIS, the original value of which is in value range $[-1, 1]$. Here, we evaluate the cosine similarity between the softmax-activated logits, or probabilities of each class, of the original image and the label. Therefore, the value exactly lies in the value range $[0, 1]$ due to the non-negative characteristic of probabilities. In addition, it also meets the demand of the second feature for MIS. The design of MIS is grounded in the simple and intuitive observation that heavy augmentations typically increase the training loss of data, which inspires us to focus on and utilize the relationship between the magnitudes and the predicted logits. Meanwhile, theoretical proofs have demonstrated the effectiveness of cosine similarity as an indicator to evaluate the contribution to model generalization \citep{theory_cos}.

In the single-label classification task, the labels are one-hot vectors, where the cosine similarity represents the predicted probability of the target class. However, inconsistency exists between classification accuracy and confidence of the prediction on image recognition tasks \citep{loss_acc_inconsistency}, especially when there are numerous target classes. The phenomenon is also shown in Fig. \ref{fig:loss}. This brings a negative effect to the cosine similarity-based MIS, where tasks with more classes generally get smaller scores compared with tasks with fewer classes. Therefore, we also introduce a hyperparameter named scaling factor $\gamma$ to normalize MIS for different tasks. The final applied MIS in this work is denoted as 
\begin{eqnarray}
	\mathrm{MIS}_i = \cos(\mathbf{p_i^{ori}}, \mathbf{l_i})^{\gamma}
        =(\frac{\mathbf{p_i^{ori}}\cdot\mathbf{l_i}}{\Vert \mathbf{p_i^{ori}}\Vert \Vert \mathbf{l_i}\Vert})^{\gamma},
\label{eq:MIS}
\end{eqnarray}
where $\cos$ denotes cosine similarity function, and $\gamma$ is the MIS scaling factor. $\mathbf{p^{ori}_{i}}$ and $\mathbf{l}_{i}$ are the prediction and the label of sample $i$. To adjust the value of $\gamma$ in different tasks, we define a new formula that is denoted as
\begin{eqnarray}\label{gamma}
	\gamma = \frac{\epsilon}{\log{c}},\, \epsilon \geq 0,
\end{eqnarray}
where $\epsilon$ is the hyperparameter to control the normalization scale of MIS, and $c$ is the number of classes in the target task. The restriction $\epsilon \geq 0$ ensures MIS lies in the range $(0,1]$. 
We choose this formula because it yields the same MIS for any $c$ when the predicted probability is uniformly distributed in each class. With this formula, tasks with more classes require smaller cosine similarity to have the same MIS compared to the ones with fewer classes, which compensates for the difficulty of these tasks to have larger confidence in the target class. 

The sample perception step generates scores to measure the difficulty of different images during training and dynamically updates these scores as training progresses. This step is expected to guide data augmentation in generating variants of original images with sample-aware magnitudes. The calculation of MIS is based on a formula that requires no additional search, avoiding the complex process to determine the data augmentation. Another advantage is that this step only requires a forward pass of the images, and the time-consuming backward update can be omitted. Therefore, it can be easily integrated into current training pipelines without obviously increasing the training time.

\textbf{Step 3: Distribution Refinement.} With the MIS calculated in Step 2, sample-aware data augmentation can be conveniently applied during model training. Since one sub-policy may contain multiple augmentation operators while there is only one MIS for each sample, the calculated MIS is shared among all operators in the same sub-policy as the magnitude. 
Apart from the magnitudes used in this step, other procedures are the same as Step 1.  

Note that although the MIS module can be applied in more augmentation frameworks that require to settle magnitudes for augmentation operators, the proposed SRA is not simply a combination of MIS and RA. The asymmetric augmentation strategy ensures the effectiveness and generalization of MIS, making the designs of SRA as a whole for out-of-the-box use. Additionally, although the exploration phase uses a random augmentation strategy that appears non-sample-aware, the refinement phase ensures all samples are utilized equally throughout training, as the training set is not split into separate subsets for exploration and refinement. Consequently, SRA inherently maintains sample-awareness.

The refinement step enables receiving real-time feedback from the target model to adjust magnitudes, sharing similar characteristics with the adaptive data augmentation in AdaAugment \citep {AdaAugment}. However, it should be noted that our refinement step requires no additional network or optimization for this feedback. Only a heuristic formula is computed, making it easily integrable into existing training pipelines.

\begin{algorithm}[t]
\small
\caption{Sample-aware RandAugment}
\label{SRA}

\KwIn{Image batches $B$ and corresponding labels $y$, training dataset $D_{train}$, augmentation depth $D$, candidate operator set $\mathcal{O}$, uniform sampling $\mathcal{U}$, model $\mathcal{M}$}
\KwOut{Trained model $\mathcal{M}$}

\For {$(B,y)$ in $D_{train}$}
    {
    Randomly split $(B,y)$ into $(B_1,y_1)$ and $(B_2,y_2)$
    
    Count the number of samples in $B_2$ as $N_2$
    
    $\color[HTML]{036400}{\mathbf{\# \; Step\; 1: \; Distribution \; Exploration}}$ 
    
    \For {$I_i$ in $B_1$}
        {Independently sample $D$ operators from $\mathcal{O}$ and $D$ magnitudes from $\mathcal{U}(0,1)$ 
        
        Augment $I_i$ to $I_i'$ with the sampled operators and magnitudes
        }
    Combine $I_i'$s to $B_1'$, and update $\mathcal{M}$ to $\mathcal{M}'$ using $B_1'$ 

    $\color[HTML]{036400}{\mathbf{\# \; Step\; 2: \; Sample \; Perception}}$
    
    Calculate the logits of $\mathcal{M}'$ for predicting $B_2$
    
    Calculate $\{MIS_j|1\leq j\leq N_2\}$ of each sample in $B_2$ using Eq. \ref{eq:MIS} 

    $\color[HTML]{036400}{\mathbf{\# \; Step\; 3: \; Distribution \; Refinement}}$

    \For {$I_j$ in $B_2$}
        {Independently sample $D$ operators from $\mathcal{O}$ 
        
        Augment $I_j$ to $I_j'$ with the sampled operators and $MIS_j$
        }
    Combine $I_j'$s to $B_2'$, and update $\mathcal{M}'$ to $\mathcal{M}$ using $B_2'$ 
    }
\end{algorithm}

\section{Experiment}\label{sec4}

We conduct several experiments to evaluate the performance of SRA on five benchmarks with different features or distributions: CIFAR-10 \citep{CIFAR}, CIFAR-100 \citep{CIFAR}, ImageNet \citep{ImageNet}, Food101\citep{Food101}, and ImageNet-LT \citep{ImageNet-LT}, and compare it with other mainstream AutoDA. The performances of other methods are from their original paper if not specially mentioned. However, we notice that the settings of different methods are not the same. For relatively fair comparisons, we also report the performance of our SRA under different settings to ensure comparability with other methods. 

Apart from benchmark comparisons, we also show the compatibility of SRA with other augmentation frameworks such as Tied Augment \citep{TiedA}. Experiments with Tied Augment assess the potential of SRA in the application of multi-view augmentation applications, which is a common paradigm in semi-supervised learning and contrastive learning. We also compare the performance of SRA with online AutoDA methods that integrate repeated augmentation \citep{BA} (i.e., Batch Augment), demonstrating its potential for compatibility with stability regularization.

The hyperparameter settings of our SRA generally follow previous works. To demonstrate the benefit of SRA to downstream tasks, we further evaluate the performance of it on object detection. All experiments are run three times, with average performance and standard deviations reported for self-implemented experiments.

\subsection{CIFAR-10 \& CIFAR-100}
Following previous works, we evaluate our SRA on two models Wide-ResNet-28-10 \citep{WRN} (WRN-28-10) and ShakeShake-26-2x96d \citep{ShakeShake} (SS-26-2x96d). Performances of Top-1 accuracy (\%) of different AutoDA methods are shown in Table \ref{CIFAR}. For a fair comparison, only methods with training epochs and tricks similar to ours are listed in the table. Note that we also label whether the method requires search in the table, where search-free methods are expected to be more convenient for wide applications. We mark RA as both search-based and search-free because it has only a few policy parameters that are easily tunable via human priors. TA has two search spaces, RA and Wide. The RA space is identical to that of SRA, while the Wide space features wider magnitude ranges for each operator. 

As a search-free method, SRA outperforms other search-free methods in many cases, while achieving competitive or even slightly better performance than search-based ones. For SS-26-2x96d on CIFAR-100, SRA is slightly worse than TA (Wide). We analyze this is because CIFAR-100 prefers wider magnitude ranges, especially where ShakeShake views more samples in the wider ranges due to the longer training period (1800 epochs). The experiments demonstrate that our SRA can improve the performance of search-free AutoDA methods under many conditions on CIFAR benchmarks.

\begin{table*}[!t]
\centering
\renewcommand\arraystretch{1.1}
\small
\captionof{table}{Accuracy (\%) on CIFAR. We label whether the listed AutoDA require search. SS: ShakeShake. \textbf{Bold}: Best performance. \underline{Underline}: Second best performance.}
\resizebox{0.8\linewidth}{!}{
\begin{tabular}{@{}lccccc@{}}
\hline
\multirow{2}{*}{Method}   & \multirow{2}{*}{\begin{tabular}[c]{@{}c@{}}Search\\ -based\end{tabular}} & \multicolumn{2}{c}{CIFAR-10}                  & \multicolumn{2}{c}{CIFAR-100}   \\ 
\cline{3-6}
  &  & WRN-28-10  & SS-26-2x96d              & WRN-28-10    & SS-26-2x96d  \\ 
\hline
AA \citep{AA}             & {$\checkmark$}     & 97.4                          & 98.0                  & 82.9                  & 85.7  \\
FastAA \citep{FastAA}     & {$\checkmark$}     & 97.3                          & 98.0                  & 82.8                  & 85.4  \\
DDAS \citep{DDAS}         & {$\checkmark$}     & 97.3                          & 98.0                  & 83.4                  & 85.0  \\
AdaAug \citep{AdaAug}     & {$\checkmark$}     & 97.4                          & -                     & 82.9                  & -  \\
TeachA \citep{TeachAug}   & {$\checkmark$}     & 97.0                          & -                  & 82.7                  & - \\
DeepAA \citep{DeepAA}     & {$\checkmark$}     & 97.56                         & 98.11                 & 84.02                 & 85.19  \\
LA3 \citep{LA3}           & {$\checkmark$}     & \underline{97.80}   & 98.07                 & \underline{84.54}   & 85.17     \\
BDA \citep{BDA}           & {$\checkmark$}     & 97.49                         & 98.05                 & 83.48                 & 85.01  \\
SLACK \citep{SLACK}       & {$\checkmark$}     & 97.46                         & -                     & 84.08                 & - \\
MADAug \citep{MADAug}     & {$\checkmark$}     & \textbf{97.9}                         & 98.2                  & 83.9                  & \underline{85.9} \\
FreeA \citep{FreeA}       & {$\checkmark$}     & 97.66                         & -                     & 84.13                 & - \\
AdaAugment \citep{AdaAugment}       & {$\checkmark$}     & 97.66                         & -                     & 83.23                 & - \\
RA \citep{RA}             & {$\checkmark$} / {\ding{55}}  & 97.3                  & 98.0     & 83.3  &  -   \\
UA \citep{UA}             & {\ding{55}}        & 97.33                         & 98.10                 & 82.82                 &  84.99   \\
TA (RA) \citep{TA}        & {\ding{55}}        & 97.46                         & -                     & 83.54                 & -  \\
TA (Wide) \citep{TA}      & {\ding{55}}        & 97.46                         & \underline{98.21}     & 84.33                 & \textbf{86.19}  \\
EntA \citep{EntAug} & \ding{55}        & 97.47                         & -                     & 83.09                 & - \\
\hline
\textbf{SRA (Ours)}                & {\ding{55}}        & 97.67 $\pm$ 0.02    & \textbf{98.36 $\pm$ 0.08}   & \textbf{84.64 $\pm$ 0.04}   & 85.74 $\pm$ 0.05\\ 
\hline
\end{tabular}
}
\label{CIFAR}
\end{table*}

\subsection{ImageNet}
To show the performance of SRA on a larger and more challenging dataset ImageNet that contains 1,000 categories and 1.3 million images, we evaluate SRA with a classical CNN model ResNet-50 \citep{resnet} and a larger one ResNet-200. We compare both Top-1 and Top-5 accuracy on this dataset with other methods. Both the performance of SRA with and without label smoothing \citep {label_smoothing} is shown in Table \ref {ImageNet}. Since RA is a key comparison focus and uses fewer training epochs (180 vs. 270) on ImageNet in its original paper, we reproduced its results under our settings for a fair comparison. 

For comparison on ResNet-50, SRA significantly outperforms all search-free methods, while also achieving better performance than many search-based ones. Our SRA is search-free, requiring only minor modifications to the traditional training pipeline and no complex search procedure to achieve improved performance. While for comparison on ResNet-200, most search-free methods do not report the results.
Our SRA outperforms the reproduced RA under both settings, meanwhile achieving competitive performance over the search-based ones. The results demonstrate that SRA shrinks the gap between the search-free and search-based AutoDA, especially on models with more parameters and deeper layers.

\begin{table*}[!t]\centering
\renewcommand\arraystretch{1.1}
\small
\captionof{table}{Top-1 and Top-5 accuracy (\%) on ImageNet of different AutoDA methods under two settings. LS: Label smoothing.}
\resizebox{0.8\linewidth}{!}{
\begin{tabular}{clcccc}
\hline
\multirow{2}{*}{LS}   & \multirow{2}{*}{Method} & \multicolumn{2}{c}{ResNet-50} & \multicolumn{2}{c}{ResNet-200} \\ 
\cline{3-6}
          &                & Top-1         & Top-5         & Top-1          & Top-5         \\ 
\hline
\multirow{9}{*}{\ding{55}}          & AA \citep{AA}            & 77.6          & 93.8          & 80.0           & 95.0          \\
& FastAA \citep{FastAA}                  & 77.6          & 93.7          & 80.6           & 95.3          \\
& DDAS \citep{DDAS}        & 78.0          & -           & 80.5        & -                              \\
& DeepAA \citep{DeepAA}                  & 78.30         & -             & \textbf{81.32} & -             \\
& BDA \citep{BDA}                     & 78.12         & 93.87         & 80.14          & 95.09         \\
& AdaAugment \citep{AdaAugment}       & 78.2         & -         & -          & -         \\
& RA \citep{RA}                      & 77.6          & 93.8          & -              & -             \\ 
& RA (repro.)              & \textbf{78.06 $\pm$ 0.01} & 93.82 $\pm$ 0.02 & 80.43 $\pm$ 0.13 & 95.16 $\pm$ 0.02 \\
& UA \citep{UA}                      & 77.63         & -             & 80.4           & -             \\
& TA (Wide) \citep{TA}                & 78.07         & 93.92         & -              & -             \\ 

\cline{2-6}
& \textbf{SRA (Ours)}               & 78.31 $\pm$ 0.09  & \textbf{94.02 $\pm$ 0.03}  & 81.11 $\pm$ 0.09  & \textbf{95.56 $\pm$ 0.02}   \\ \hline
\multirow{3}{*}{$\checkmark$} & TeachA \citep{TeachAug} & 77.8         & 93.7         & -              & - \\ 

& LA3 \citep{LA3}                     & 78.71         & -             & -              & -             \\
& RA (repro.)              & 78.53 $\pm$ 0.04  & 94.20 $\pm$ 0.01 & 81.00 $\pm$ 0.04 & 95.32 $\pm$ 0.02 \\ 
\cline{2-6}
& \textbf{SRA (Ours)}               & \textbf{78.83 $\pm$ 0.07}  & \textbf{94.24 $\pm$ 0.03}  & \textbf{81.70 $\pm$ 0.05}   & \textbf{95.79 $\pm$ 0.04}  \\ 
\hline
\end{tabular}
}
\label{ImageNet}
\end{table*}

To further evaluate the generalization of SRA on different neural architectures, we also compare the performance of SRA using vision Transformer DeiT-Tiny \citep{DeiT} without distillation, Swin-Tiny \citep{Swin}, and VMamba-Tiny \citep{VMamba}. The results are shown in Table \ref{DeiT_Swin_VMamba}. For all these experiments, augmentations other than basic transformation and SRA are not applied to reduce the impact of joint usage of different augmentations. Note that RA with a standard deviation for magnitude (proposed in timm \citep {timm}) is also adopted in the original implementations of DeiT, Swin, and VMamba. In addition, we report the results of SRA both without and with scaling factor $\gamma$ to show its advantage for direct application to new neural architectures and also the performance gain from hyperparameter tuning.

SRA sightly outperforms the comparing DA methods on Top-1 accuracy in all these architectures even without tuning scaling factor $\gamma$, further demonstrating the potential of it for wide applications on different types of neural networks in vision classification. After tuning $\gamma$, where we set $\epsilon=8$ for DeiT and $\epsilon=1$ for Swin and VMamba, the Top-1 performance can be further improved. We note that the Top-5 performance of Swin and VMamba is nearly unchanged. This indicates that the scaling factor may work through strengthening the confidence on the correct class when using these models. We also notice performance gaps between RA (2, 9/0.5) and original performance, which rise from different augmentation settings.

\begin{table}[!t]
\centering
\renewcommand\arraystretch{1.1}
\small 
\captionof{table}{Accuracy (\%) on ImageNet using architectures apart from CNNs. Input image resolution is set to $224\times224$. RA (2, 9/0.5): RandAugment with $N$ of 2, $M$ of 9, and the standard deviation of $M$ of 0.5. 
*: Reimplemented results without extra augmentations.}
\resizebox{1.0\linewidth}{!}{
\begin{tabular}{c|lcc}
\hline
Model                     & Method               & Top-1                & Top-5                \\ 
\hline
\multirow{5}{*}{DeiT-T}   
                          & Basic                & 69.30 $\pm$ 0.02     & 88.34 $\pm$ 0.01     \\
                          & RA (2, 9)            & 73.76 $\pm$ 0.07     & 91.41 $\pm$ 0.06     \\
                          & RA (2, 9/0.5)*        & 73.84 $\pm$ 0.13     & 91.37 $\pm$ 0.03     \\ \cline{2-4} 
                          & \textbf{SRA (w/o $\gamma$)}  & 73.90 $\pm$ 0.14     & 91.39 $\pm$ 0.03     \\ 
                          & \textbf{SRA (w/ $\gamma$)}  & \textbf{74.06 $\pm$ 0.10}     & \textbf{91.52 $\pm$ 0.09}     \\ 
\hline
\multirow{5}{*}{Swin-T}   
                          & Basic                & 74.71$\pm$ 0.17      & 91.60 $\pm$ 0.13     \\
                          & RA (2, 9)            & 78.83 $\pm$ 0.05     & 93.83 $\pm$ 0.01     \\
                          & RA (2, 9/0.5)*        & 79.08 $\pm$ 0.04     & 93.97 $\pm$ 0.02     \\ \cline{2-4} 
                          & \textbf{SRA (w/o $\gamma$)}  & 79.13 $\pm$ 0.12     & 94.12 $\pm$ 0.02     \\ 
                          & \textbf{SRA (w/ $\gamma$)}  & \textbf{79.31 $\pm$ 0.03}     & \textbf{94.15 $\pm$ 0.04}     \\ 
\hline
\multirow{5}{*}{VMamba-T} 
                          & Basic                & 77.08 $\pm$ 0.16     & 92.81 $\pm$ 0.05     \\
                          & RA (2, 9)            & 80.28 $\pm$ 0.07     & 94.49 $\pm$ 0.03     \\
                          & RA (2, 9/0.5)*        & 80.59 $\pm$ 0.01     & 94.77 $\pm$ 0.04     \\ \cline{2-4} 
                          & \textbf{SRA (w/o $\gamma$)}  & 80.63 $\pm$ 0.06     & 94.80 $\pm$ 0.06     \\ 
                          & \textbf{SRA (w/ $\gamma$)}  & \textbf{80.89 $\pm$ 0.04}     & \textbf{94.81 $\pm$ 0.04}     \\ 
\hline
\end{tabular}
}
\label{DeiT_Swin_VMamba}
\end{table}

\subsection{Food101}

To demonstrate the generalization ability of SRA, we also conduct experiments on benchmarks with different data distributions. We first evaluate SRA on a more fine-grained image recognition benchmark Food101 \citep{Food101}. The dataset contains food of 101 categories, with 750 images for training and 250 for validation per class. We treat this benchmark as a new task distinct from classical benchmarks, and adopt training configurations from ImageNet of ResNet-50 with label smoothing, except that no scaling factor $\gamma$ is applied in SRA to simulate the situation with limited prior for new tasks. We compare the performance of ResNet-50 under different augmentation settings (basic augmentation, RA with the best policy found on ImageNet, and SRA without $\gamma$) using the transferred configurations. Top-1 accuracy on the dataset is shown in Table \ref{food101}. RA results are reproduced under our settings.

As shown, SRA significantly outperforms the models under the settings of basic augmentation and RA (2, 9), indicating a better generalization ability of SRA in new tasks. The results also show the potential of SRA to shorten the adaption time and cost for AutoDA in new applications, which is valuable for the fast development of community.

\begin{table}[!h]
\centering
\small
\renewcommand\arraystretch{1.1}
\captionof{table}{Top-1 accuracy (\%) on fine-grained classification benchmark Food101.} 
\resizebox{1.0\linewidth}{!}{
\begin{tabular}{lccc}
\hline
            & Basic             & RA (2, 9)       & \textbf{SRA (w/o $\gamma$)}       \\ 
\hline
ResNet-50   & 83.18 $\pm$ 0.06    & 85.97 $\pm$ 0.03  & \textbf{87.04 $\pm$ 0.02} \\ 
\hline
\end{tabular}
}
\label{food101}
\end{table}

\subsection{ImageNet-LT}

To show the generalization ability of SRA on data with imbalanced distribution, we further evaluate SRA on ImageNet-LT \citep{ImageNet-LT}, which is a long-tail subset of the ImageNet. ImageNet-LT includes classes with varying numbers of images, which requires SRA to improve model performance on recognizing infrequent classes while maintaining overall accuracy. Similar to experiments on Food101, we compare the performance of ResNet-50 under different augmentation settings using the transferred configurations. Top-1 accuracy on the whole dataset and categories with different number of samples is shown in Table \ref{ImageNet-LT}. SRA outperforms RA with a 1.58\% accuracy improvement under the same training configuration. The performance of classes with different shots is also improved across the board, demonstrating the strength of SRA in image recognition across benchmarks with imbalanced data distribution.

\begin{table*}[!h]
\centering
\small 
\renewcommand\arraystretch{1.1}
\captionof{table}{Top-1 accuracy (\%) on long-tailed benchmark ImageNet-LT. Numbers under different shots show the number of training samples in one category.} 
\resizebox{0.8\linewidth}{!}{
\begin{tabular}{l|ccc|c}
\hline
\begin{tabular}[c]{@{}l@{}}Backbone\\ ResNet-50 (\%)\end{tabular} & \begin{tabular}[c]{@{}c@{}}Many-shot\\ \textgreater{}100\end{tabular} & \begin{tabular}[c]{@{}c@{}}Medium-shot\\ $\leq$ 100 \& \textgreater 20\end{tabular} & \begin{tabular}[c]{@{}c@{}}Few-shot\\ $\leq$ 20\end{tabular} & Overall                 \\ \hline
Basic                                                             & 58.38 $\pm$ 0.31                                                        & 32.79 $\pm$ 0.26                                                                      & 11.64 $\pm$ 0.15                                               & 39.71 $\pm$ 0.25          \\
RA (2, 9)                                                         & 63.75 $\pm$ 0.16                                                        & 38.97 $\pm$ 0.06                                                                      & 15.12 $\pm$ 0.22                                               & 45.18 $\pm$ 0.08          \\ \hline
\textbf{SRA (w/o $\gamma$)}                                       & \textbf{65.22 $\pm$ 0.23}                                               & \textbf{40.75 $\pm$ 0.10}                                                             & \textbf{16.41 $\pm$ 0.35}                                      & \textbf{46.76 $\pm$ 0.09} \\ \hline
\end{tabular}
}
\label{ImageNet-LT}
\end{table*}

\begin{figure*}[!h]

	\centering
    \includegraphics[width=0.75\linewidth]{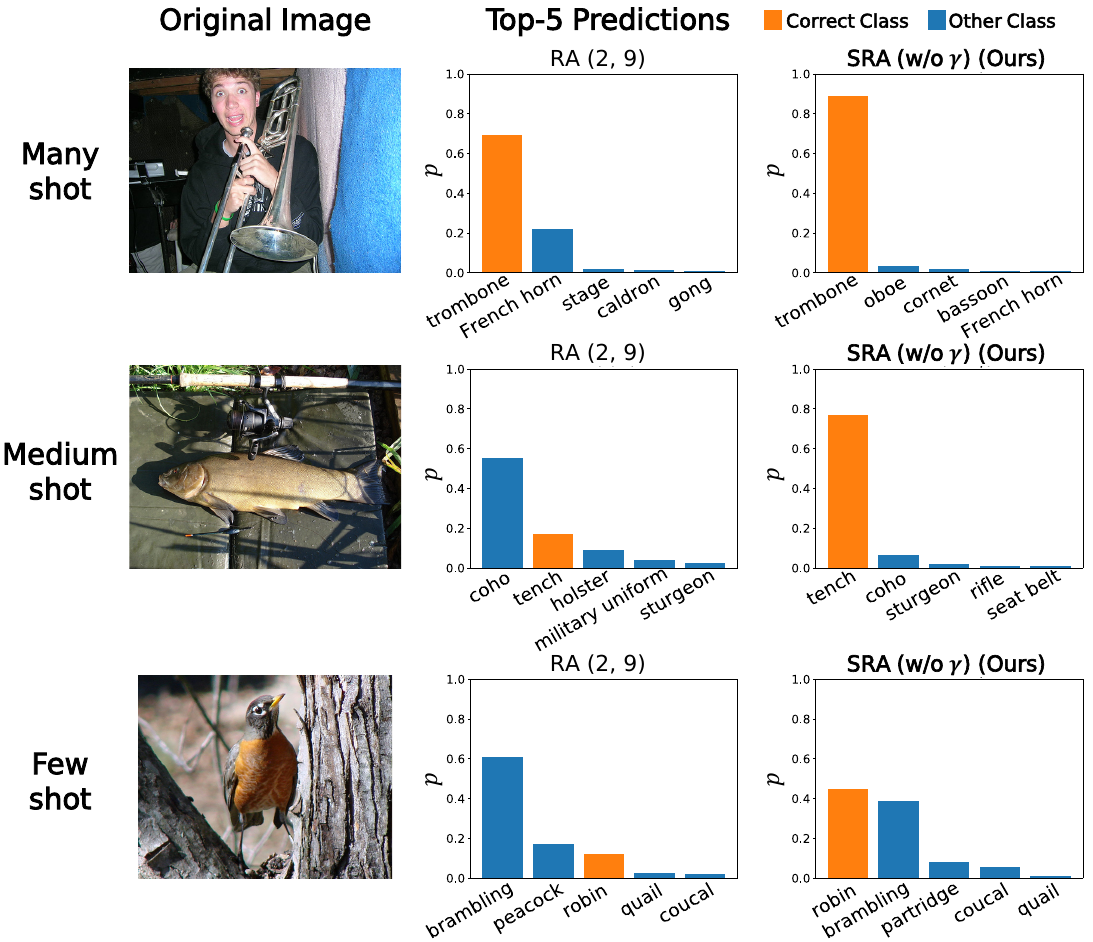}
	\captionof{figure}{Representative cases of categories with different shots in ImageNet-LT that show how SRA improves image recognition ability compared with RA.}
	\label{ImageNet-LT-visual}
\end{figure*}

We further conduct a qualitative analysis of how SRA improves image recognition compared to RA on ImageNet-LT via visualization, which is shown in Fig. \ref{ImageNet-LT-visual}. For many-shot classes, SRA enhances confidence in the correct class. For medium-shot classes, SRA helps distinguish objects from fine-grained classes. For few-shot classes, SRA smooths the predictions and boosts confidence in the correct class.

\subsection{Combination with Tied Augment}

Tied Augment \citep{TiedA} is a recently proposed augmentation framework that is inspired by contrastive learning. It uses multiple views of the same batch, which significantly improves the representation ability of models with the alignment of features and logits of different views. Since Tied Augment requires combination with other data augmentation methods, we integrate SRA with it to evaluate compatibility and performance, denoted as Tied-SRA. For a fair comparison, we also reproduce the Tied Augment with RA (Tied-RA) under the same training settings of SRA. The results are shown in Table \ref{TiedA}. Tied-SRA outperforms reproduced Tied-RA on both CIFAR benchmarks, illustrating the advantage of SRA on aligning different views of the original images. However, we note that the reproduced Tied-RA results are slightly worse than the original ones, which may arise from the lack of detailed configurations of the training hyperparameters.

\begin{table}[!h]
\centering
\renewcommand\arraystretch{1.1}
\small 
\captionof{table}{Accuracy (\%) on CIFAR with Tied Augment.}
\resizebox{0.9\linewidth}{!}{
\begin{tabular}{lcc}
\hline
Methods          & CIFAR-10         & CIFAR-100                 \\ \hline
Tied-RA          & \textbf{98.1}    & 85.0                      \\
Tied-RA (repro.) & 97.89 $\pm$ 0.07 & 84.80 $\pm$ 0.26          \\ \hline
\textbf{Tied-SRA (Ours)}  & 98.04 $\pm$ 0.05 & \textbf{85.43 $\pm$ 0.14} \\ \hline
\end{tabular}
}
\label{TiedA}
\end{table}


\subsection{Combination with Batch Augment}

Many online AutoDA methods that dynamically adapt augmentation policy during training are combined with Batch Augment \citep{BA} (BA) that repeatedly augment the same batch into different variants, aiming at estimating the loss expectations of the augmented images for learning optimal policy parameters. These methods show superior performance compared to those without BA, but the duplication of augmentation requires more time and GPU memories for training. To compare our SRA with online methods, we combine SRA with BA and report the performance. We also display the results of the state-of-the-art search-free method TA (Wide) repeated by 8 times as a reference. Since BA allows faster convergence compared with the training without BA \citep{BA}, we reduce the training epochs of our SRA to 35 (referring to DeepAA \citep{DeepAA}). The results are shown in Table \ref{BA}. We label the repeated times after each method, which is denoted as $\times4$ or $\times8$. 

Interestingly, we find SRA outperforms both online search-based AutoDA and search-free method TA on CIFAR. We emphasize that SRA achieves state-of-the-art performance with only 200 epochs on SS-26-2x96d under $\times8$ settings, which saves $\sim$3 times training cost compared with other methods that at least require 600 epochs (TA). The augmented images generated by BA estimate the loss expectations of the augmented images generated by the combined AutoDA. SRA shows better recognition performance through learning from the loss expectations of the augmented samples compared with previous search-free methods, indicating the augmented image distribution of SRA is closer to the target data distribution. The result demonstrates that SRA indeed improves the generalization of represented classes.

\begin{table}[!h]
\centering
\renewcommand\arraystretch{1.1}
\small 
\captionof{table}{Accuracy (\%) on CIFAR with Batch Augment. WRN: WideResNet-28-10. SS: ShakeShake-26-2x96d.}
\resizebox{0.9\linewidth}{!}{
\begin{tabular}{lcccc}
\hline
\multirow{2}{*}{Methods} & \multicolumn{2}{c}{CIFAR-10}            & \multicolumn{2}{c}{CIFAR-100}           \\ \cline{2-5} 
                         & WRN                       & SS          & WRN                       & SS \\ \hline
MetaA ($\times$4) \citep{MetaA}       & 97.76                     & 98.29       & 83.79                     & 85.97       \\
AdvAA ($\times$8) \citep{AdvAA}       & 98.10                     & 98.15       & 84.51                     & 85.90       \\ 
LatentA ($\times$8) \citep{LatentA}   & 98.16                     & -           & -                         & -           \\
TA (Wide) ($\times$8) \citep{TA}      & 98.04                     & 98.12       & 84.62                     & 86.02 \\
\hline
\textbf{SRA ($\times$8) (Ours)}                & \textbf{98.34} & \textbf{98.38} & \textbf{85.23} & \textbf{86.65}  \\ \hline
\end{tabular}
}
\label{BA}
\end{table}

\subsection{Performance on Object Detection}

We further evaluate the performance of SRA in the downstream task object detection. We train RetinaNet \citep {retinanet} with a ResNet-50 backbone pretrained using RA and SRA, respectively, and evaluate the performance using COCO \citep{COCO}. Results are in Table \ref{detection}. Training configurations and the baseline (Basic configuration in the table) are from mmdetection\footnote{https://github.com/open-mmlab/mmdetection/tree/main} for "R-50-FPN 2x". Each configuration is trained three times, with different pretrained weights for each run.

SRA-pretrained backbones can improve the performance of object detection in both overlapped areas and different sizes of the target bboxes. Although the recognition ability of SRA is slightly worse than RA for larger objects, the recognition precision and overall performance are better. The results show that SRA can improve the image recognition ability in tasks other than classification.

\begin{table}[!h]
\centering
\renewcommand\arraystretch{1.1}
\small
\captionof{table}{Object detection performance on COCO.}
\resizebox{\linewidth}{!}{
\begin{tabular}{lcccccc}
\hline
Augment     & $AP$            & $AP_{50}$       & $AP_{75}$       & $AP_{S}$        & $AP_{M}$        & $AP_{L}$     \\ 
\hline
Basic       & 37.4          & 56.7          & 39.6          & 20.0          & 40.7          & 49.7          \\
RA (repro.) & 38.0          & 57.3          & 40.4          & \textbf{21.3} & 41.3          & \textbf{50.7} \\ 
\hline
\textbf{SRA (Ours)}  & \textbf{38.3} & \textbf{57.7} & \textbf{41.0} & \textbf{21.3} & \textbf{41.9} & 50.5          \\ 
\hline
\end{tabular}
}
\label{detection}
\end{table}

\section{Discussion}

\subsection{Impact of Normalization Scale $\epsilon$} 

\begin{figure}[!h]
\centering
\includegraphics[width=1.0\linewidth]{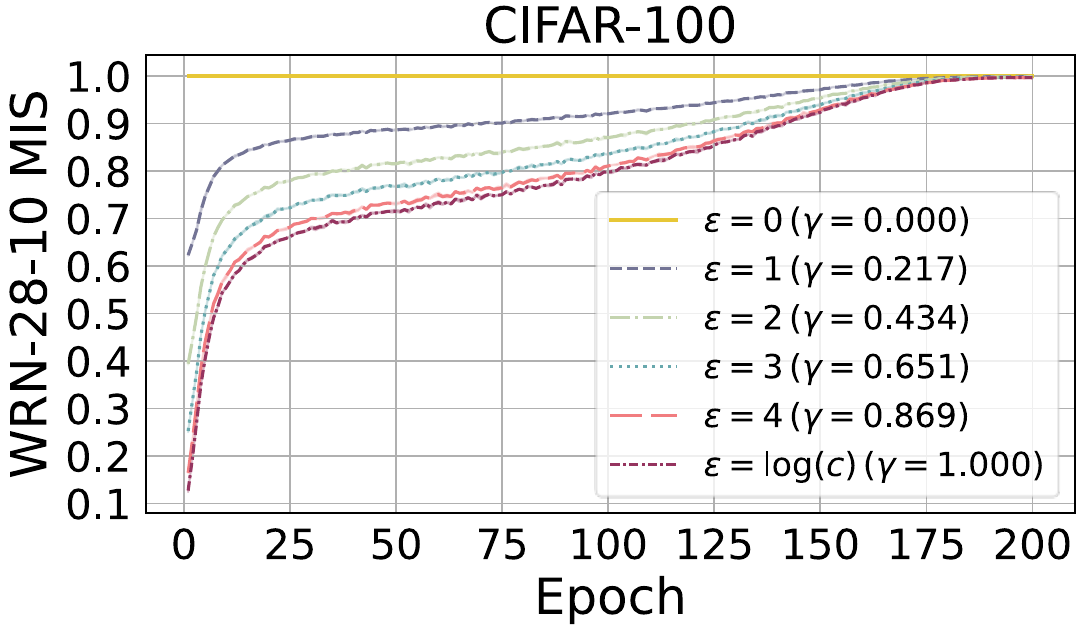}
\captionof{figure}{Curves for the dynamic changes of the average MIS of SRA during training.}
\label{Fig: e-epoch}
\end{figure}

We evaluate the impact of the scaling factor $\epsilon$ on CIFAR-100 using WRN-28-10, the results of which are shown in Table \ref{Impact of epsilon}. Meanwhile, we also show how the MIS changes during training in Fig. \ref{Fig: e-epoch}. Training with a larger $\epsilon$ yields steeper MIS curve, indicating faster changes in the distribution of augmented samples. However, the performance does not increase monotonically with increasing $\epsilon$. This phenomenon indicates that adjusting the data distribution requires balancing convergence speed and sample diversity.

\begin{table}[!h]
\centering
\renewcommand\arraystretch{1.1}
\small 
\captionof{table}{Impact of different values of the scaling factor $\epsilon$ on model performance using WRN-28-10. Results are evaluated on CIFAR-100.}
\resizebox{\linewidth}{!}{
\begin{tabular}{lccc}
\hline
$\epsilon$                     & 0              & 1              & 2                        \\ 
\hline
Accuracy (\%)                  & 84.44 $\pm$ 0.12 & 84.48 $\pm$ 0.14 & \textbf{84.64 $\pm$ 0.04}   \\ 
\hline \hline
\multirow{2}{*}{(continue)}    & 3              & 4              & $\log c$  \\ 
\cline{2-4}
                               & 84.49 $\pm$ 0.17 & 84.57 $\pm$ 0.18 & 84.57 $\pm$ 0.16 \\ 
\hline
\end{tabular}
}
\label{Impact of epsilon}
\end{table}

\subsection{Impact of Augmentation Depth $D$} 

Augmentation depth is an important hyperparameter that determines the scale of the search space. With multiple operators sequentially applied, the diversity of the augmented data increases, but the data also becomes more challenging to learn from. Following previous settings in AA and RA, we primarily select augmentation depth as 2 in our experiments. To evaluate the impact of this hyperparameter on the proposed SRA, we conduct experiments on CIFAR-100 using WRN-28-10. As shown in Table \ref{aug_depth}, the balance between diversity and the representation ability of the target model is necessary for SRA. Augmentation depth $D=2$ performs the best, which is consistent with the results of previous work RA \citep{RA}.

\begin{table*}[!h]
\centering
\small 
\renewcommand\arraystretch{1.1}
\captionof{table}{Impact of augmentation depth $D$ on model performance using WRN-28-10. Results are evaluated on CIFAR-100.}
\resizebox{0.75\linewidth}{!}{
\begin{tabular}{lcccc}
\hline
$D$                            & 1                & 2                           & 3                & 4     \\
\hline
Accuracy (\%)                  & 84.49 $\pm$ 0.08 & \textbf{84.64 $\pm$ 0.04} & 84.56 $\pm$ 0.16 & 84.02 $\pm$ 0.11 \\
\hline
\end{tabular}
}
\label{aug_depth}
\end{table*}

\subsection{Impact of Augmentation Operators}

Since operators are significant for generating augmented images, it is necessary to evaluate the impact of different operators for AutoDA. In SRA, we simply adopt the candidate operator set from RA \citep{RA} with random sampling to reduce the complexity of the method. However, it is obvious that different operators contribute diversely to the model performance. As a result, we performed operator ablation experiments on CIFAR using WRN-28-10 in Table \ref{aug_op} to show the sensitivity of SRA to each operator in the search space. For each line, we delete one specific operator from the original 14 operators, and report the performance of SRA on the remaining 13 operators. Each experiment is run for 3 times. As shown, SRA is sensitive to the selection of augmentation operators, which is also one of the characteristics of previous AutoDA \citep{RA, DADA, DeepAA}. The results indicate that current design of the selection of operators in SRA is suboptimal.

\begin{table}[!h]
\centering
\small
\renewcommand\arraystretch{1.1}
\captionof{table}{Impact of augmentation operators on CIFAR-10 using WRN-28-10. Operators that show negative effect are shown with underline.}
\resizebox{1.0\linewidth}{!}{
\begin{tabular}{lcccc}
\hline
\multirow{2}{*}{\begin{tabular}[c]{@{}l@{}}Ablated\\ Operator\end{tabular}} & \multicolumn{2}{c}{CIFAR-10} & \multicolumn{2}{c}{CIFAR-100} \\
\cline{2-5} 
                            & Accuracy   & Gain  & Accuracy    & Gain    \\
\hline
ShearX           & 97.78 $\pm$ 0.02   & \underline{-0.11}             & 84.68 $\pm$ 0.07   & \underline{-0.04} \\ 
ShearY           & 97.56 $\pm$ 0.05   & 0.11                          & 84.65 $\pm$ 0.07   & \underline{-0.01} \\ 
TranslateX       & 97.70 $\pm$ 0.01   & \underline{-0.03}             & 84.63 $\pm$ 0.14   & 0.01 \\ 
TranslateY       & 97.60 $\pm$ 0.07   & 0.07                          & 84.11 $\pm$ 0.13   & 0.53 \\ 
Rotate           & 97.58 $\pm$ 0.05   & 0.09                          & 84.39 $\pm$ 0.05   & 0.25 \\ 
Brightness       & 97.74 $\pm$ 0.06   & \underline{-0.07}             & 84.57 $\pm$ 0.14   & 0.07 \\ 
Color            & 97.59 $\pm$ 0.08   & 0.08                          & 84.54 $\pm$ 0.16   & 0.10 \\  
Sharpness        & 97.58 $\pm$ 0.03   & 0.09                          & 84.63 $\pm$ 0.20   & 0.01 \\
Contrast         & 97.65 $\pm$ 0.06   & 0.02                          & 84.46 $\pm$ 0.15   & 0.18 \\
Solarize         & 97.60 $\pm$ 0.11   & 0.07                          & 84.07 $\pm$ 0.24   & 0.57 \\
Posterize        & 97.51 $\pm$ 0.05   & 0.16                          & 84.63 $\pm$ 0.22   & 0.01 \\
Equalize         & 97.74 $\pm$ 0.01   & \underline{-0.07}             & 84.27 $\pm$ 0.23   & 0.37 \\
Autocontrast     & 97.59 $\pm$ 0.02   & 0.08                          & 84.42 $\pm$ 0.18   & 0.22 \\
Identity         & 97.61 $\pm$ 0.04   & 0.06                          & 84.77 $\pm$ 0.11   & \underline{-0.13} \\
\hline
\end{tabular}
}

\label{aug_op}
\end{table}

\subsection{Time Cost}

\begin{figure*}[!h]
\centering
\includegraphics[width=0.9\linewidth]{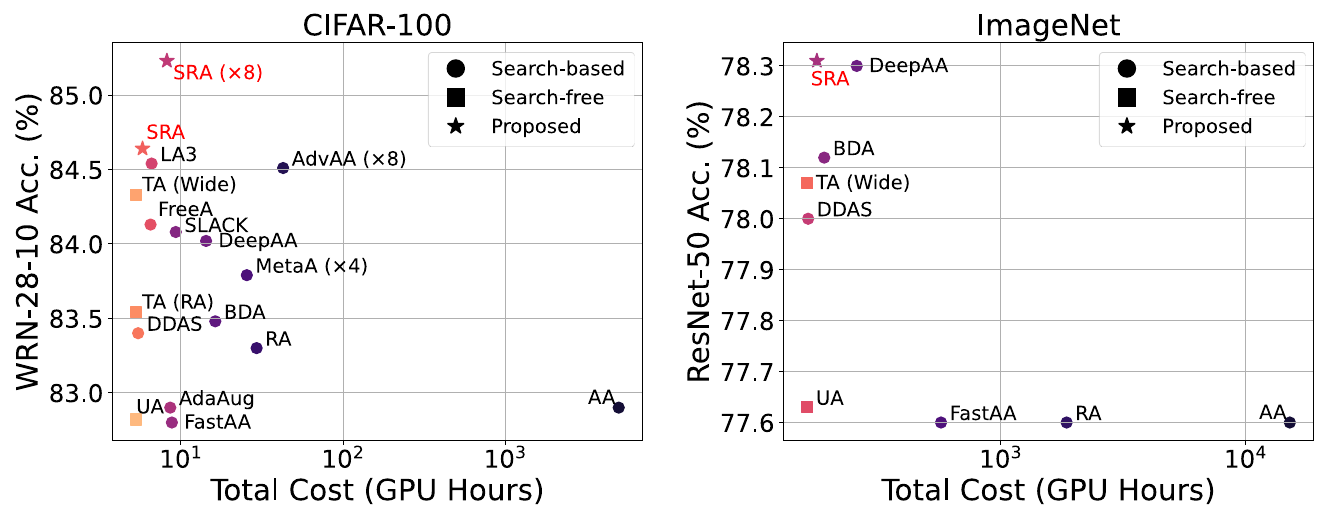}
\captionof{figure}{Total search and training cost (GPU hours) of different AutoDA methods.}
\label{Fig: search cost}
\end{figure*}

Step 2 in SRA requires an extra inference to calculate the MIS of a sub-batch compared with the traditional training pipeline. Thus, the proposed SRA requires $\times$0.5 extra forward calculation in total. This extra cost is practically cheap because back-propagation, rather than forward inference, consumes most of the training time. To evaluate the efficiency of SRA, we also report the practical extra training cost using NVIDIA GeForce RTX 3090 GPUs for CIFAR-100 and A100 for ImageNet. We find that SRA takes 105 s/epoch to training WRN-28-10 on a single GPU on CIFAR-100, and 591 s/epoch to train ResNet-50 on 4 GPUs on ImageNet. In comparison, the traditional training pipeline with augmentations such as RA takes 96 s/epoch and 540 s/epoch for CIFAR-100 and ImageNet, respectively. The training time per epoch for SRA is only $\sim$1.1 times of the traditional training pipeline, indicating its efficiency compared to many search-based AutoDA methods. With the estimated time cost on the sum of search and training time, we draw the scatter plot of performance (\% accuracy) and total training cost (GPU hours) of other AutoDA methods in Fig. \ref{Fig: search cost}. The total cost is evaluated according to the reported search cost and training epochs in the corresponding papers. As a search-free method, SRA is located in the top-left corner of this figure. It strikes a good balance between performance and time cost.

\subsection{Ablation Studies}

To better understand the characteristics of SRA, we conduct seven ablation studies: 1) The MIS scaling factor $\gamma$; 2) Random augmentation for exploring distributions; 3) The distribution exploration process; 4) The sample perception and distribution refinement process; 5) Asymmetric training strategy; and 6 \& 7) Similarity formulas for calculating MIS without $\gamma$. We separately remove or modify each part of these designs, and train the models from scratch three times. The results are shown in Table \ref{ablation}. SRA outperforms all the ablated or modified settings on CIFAR, indicating the effectiveness of the proposed design. We note that both distribution exploration and refinement significantly contribute to the performance, especially on CIFAR-100 that is more challenging. We analyze that the effectiveness of SRA is due to the positive regularization effect from hard samples significantly overwhelming the negative overfitting effect. The proposed asymmetric training strategy also contributes to the performance. Besides, more formulas to calculate MIS are worth trying, while we find cosine similarity is both intuitive and sufficiently effective for distribution refinement. The results underscore the merits of the asymmetric augmentation strategy in SRA, which may serve as a catalyst for advancing future designs of AutoDA methods.

\begin{table*}[!h]
\centering
\renewcommand\arraystretch{1.1}
\small
\captionof{table}{Ablation studies of SRA performance on CIFAR (\%).}
\resizebox{0.8\linewidth}{!}{
\begin{tabular}{lcc}
\hline
Description                                            & CIFAR-10                & CIFAR-100      \\ 
\hline
1) Remove $\gamma$                                     & 97.60 $\pm$ 0.12          & 84.49 $\pm$ 0.05          \\
\hline
2) Remove random augment in Step 1                     & 97.41 $\pm$ 0.07          & 84.45 $\pm$ 0.01          \\
3) Replace Step 1 with Step 2 \& 3                     & 97.60 $\pm$ 0.04          & 84.09 $\pm$ 0.01          \\ 
4) Replace Step 2 \& 3 with Step 1                     & 97.47 $\pm$ 0.09          & 83.93 $\pm$ 0.09          \\
5) Use both Step 1 and Step 2 \& 3 in one sub-batch               & 97.44 $\pm$ 0.05          & 84.54 $\pm$ 0.02         \\
\hline
6) Use Euclidean distance for MIS                      & 97.65 $\pm$ 0.08          & 84.37 $\pm$ 0.01          \\ 
7) Use Jaccard similarity for MIS                      & 97.56 $\pm$ 0.04          & 84.03 $\pm$ 0.13          \\
\hline
The proposed SRA                                       & \textbf{97.67 $\pm$ 0.02} & \textbf{84.64 $\pm$ 0.04} \\ 
\hline
\end{tabular}
}
\label{ablation}
\end{table*}

\subsection{Represented Feature Distribution}
\begin{figure*}[!h]
    \small
	\centering
	\includegraphics[width=0.9\linewidth]{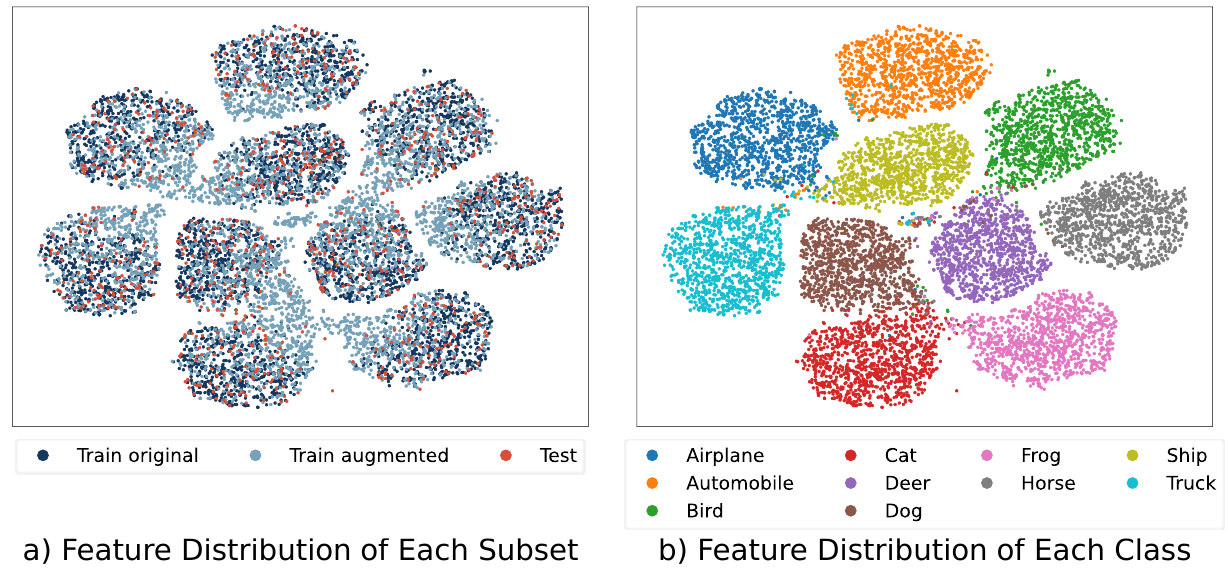}
	\captionof{figure}{Feature distributions of CIFAR-10 with SRA. Represented through well-trained WRN-28-10.}
	\label{fig:schema}
\end{figure*}

We also draw the represented data distribution of SRA in Fig. \ref{fig:schema}, where augmented data are generated using random sub-policy with corresponding MIS. Features after Global Average Pooling are first reduced to 64 dimensions using principal component analysis, and then shown in 2D t-SNE \citep{t-SNE}. The figure shows the distributions of the represented features of 10\% stratified sampled CIFAR-10 data. The augmented data generally lie at the boundaries of each cluster, indicating the effectiveness of MIS in generating hard samples. Although some augmented samples lie in other clusters due to augment ambiguity \citep{AutoDA_KL}, or are outliers due to over-transformation, they generally benefit the representation of unseen samples.

\subsection{Loss Visualization}
\begin{figure*}[!h]
	\centering
    \includegraphics[width=0.9\linewidth]{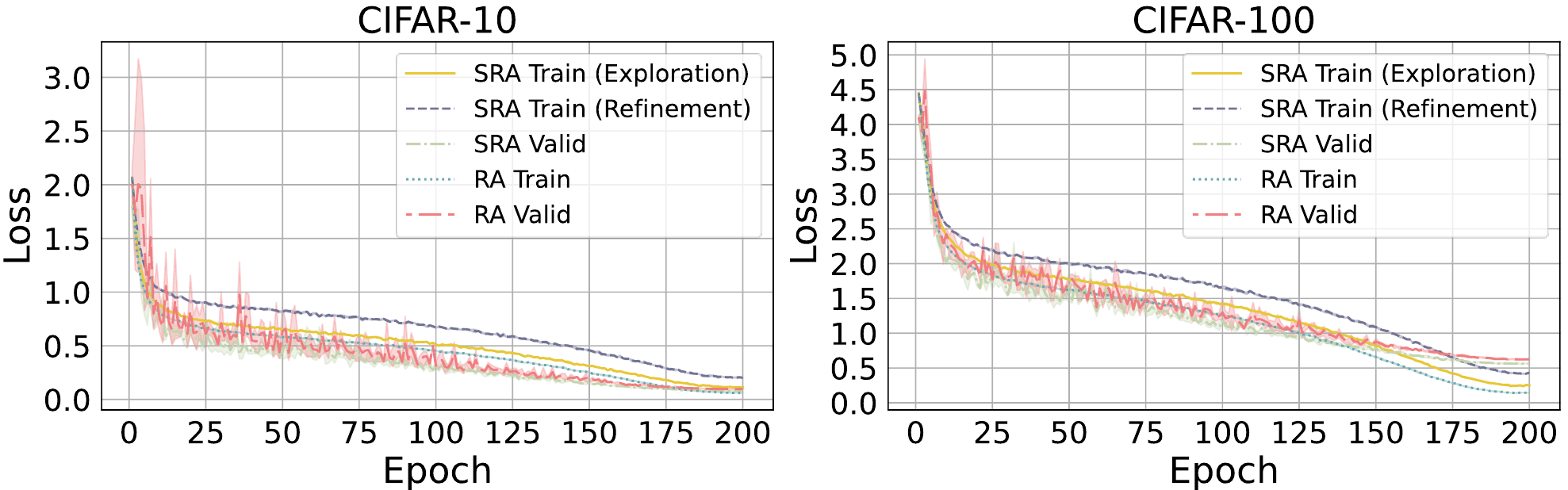}
	\captionof{figure}{The average loss curves for RA and SRA on CIFAR-10/100 using WRN-28-10. The ranges indicate the maximum and minimum loss on each epoch.}
	\label{fig:loss}
\end{figure*}

We also present the loss curves for SRA and the reproduced RA on CIFAR-10/100 to illustrate the impact of SRA on the learning process. As shown in Fig. \ref{fig:loss}, the validation loss of SRA exhibits less oscillation compared to RA, potentially due to the smoother representation of data features. Additionally, SRA shows higher training losses but lower test losses relative to that of RA, indicating improved generalization to unseen data. The distribution refinement step incurs larger losses compared to the distribution exploration step, further demonstrating its effectiveness in focusing on hard samples. However, the increase in training loss deserves attention, as it may lead to slower training convergence, thereby limiting performance under the same training budget. Moreover, this could increase the difficulty for the target model in accurately representing different classes.

\subsection{Limitations}
Although SRA shrinks the gap between search-based and search-free AutoDA, it has a small search space that limits the performance upper bound. We adopt this design considering both simplicity and efficiency. A prospective way that is beneficial for training is introducing new parameters for selecting candidate operators to learning the importance of each operator. By adopting optimization strategies to select beneficial operators, SRA may be further improved. However, the most important characteristic of SRA is its simplicity for wide application. The optimization of operator selection, which generally uses reinforcement learning, evolutionary algorithms, or gradient optimization, will undoubtedly increase the complexity of SRA, which sets barriers for tuning the proper hyperparameters and realize the implementation in other tasks. Another way is removing operators from the candidate operator set, or adding new ones, as shown in Table \ref{aug_op}. Nevertheless, for simplicity and relatively fair comparisons with previous works, we adopt the candidate operator set of RA, which is enough to demonstrate the value of our SRA. These designs are worth trying in future works.

Moreover, the hyperparameter scaling factor $\gamma$ in SRA requires tuning to achieve optimal performance. Given that it is the only hyperparameter after fixing the search space, the associated tuning cost remains relatively low, particularly when leveraging proxy tasks. Additionally, empirical results suggest that reasonable priors can often yield competitive performance even without careful tuning. In future work, we plan to explore hyperparameter optimization techniques to enable the automated and dynamic adjustment of scaling factor $\gamma$, which may further enhance model performance.

Another problem is that mainstream AutoDA methods can hardly avoid over-transformation, and neither can SRA, due to the lack of further evaluation of the semantics in augmented samples. Combining it with out-of-distribution checking may alleviate the over-transformation problem \citep{DualAug}. Besides, SRA also requires exploration in tasks beyond supervised image recognition.

\section{Conclusion}
In this work, we propose Sample-aware RandAugment (SRA), a search-free AutoDA method, to enhance the generalization ability of neural networks. The results demonstrate that heuristic designs can achieve competitive performance to optimized ones, while keeping simplicity for seamless implementation in wide applications. The proposed MIS and asymmetric augmentation strategy may inspire future works to design simpler, more effective, and more practical AutoDA, which will further contribute to the development of the community.

\bibliographystyle{spbasic}
\footnotesize
\bibliography{SRA}

\newpage
\appendix

\renewcommand{\thetable}{A\arabic{table}}
\renewcommand{\thefigure}{A\arabic{figure}}
\renewcommand{\thesection}{A\arabic{section}}
\setcounter{table}{0}
\setcounter{figure}{0}
\onecolumn

\section{Implementation Details}

For experiments on CIFAR, we apply SRA after basic cropping and random horizontal flipping, while before Cutout. For experiments on ImageNet, we only sequentially apply basic cropping, random horizontal flipping, and our SRA. Augmentation depth $D$ for all SRA experiments is set to 2. Other hyperparameters for experiments on CIFAR and ImageNet are listed in Table \ref{hyperparameter-setting}. We also apply a cosine annealing learning rate schedule with a minimum learning rate of 0 for all experiments, which is adjusted after each iteration for updating model weights.

For experiments on Food101 and ImageNet-LT, we directly apply the hyperparameters used to train the model on ImageNet to showcase the performance of different AutoDA methods when dealing with new tasks.

For experiments that combine SRA with Tied Augment, the hyperparameter settings are the same as those in CIFAR experiments. The weight to calculate the alignment loss between features of the two views is set to 20 for all experiments, which is the same as the original paper \citep{TiedA}.

For experiments that combine SRA with Batch Augment, the hyperparameters are listed in Table \ref{BA-hyperparameter-setting}.

All the experiments on CIFAR are conducted on RTX 3090 GPU, while those on ImageNet are conducted on A100. Performances are reported with three different random seeds.

\begin{table*}[!h]
\centering
\small
\renewcommand\arraystretch{1.1}
\captionof{table}{The hyperparameters for SRA in the comparing experiments with mainstream AutoDA. *: SRA adopts a batch split strategy to update model weights twice with sub-batches respectively, where batch size is required to be twice the comparing methods for a fair comparison. The practical batch size to update model weight is identical to comparing methods regardless of the boundary conditions. WE: Warmup epochs. LS: Label smoothing. WRN: WideResNet-28-10. SS: ShakeShake-26-2x96d. ResN: ResNet. $c$: The number of classes of ImageNet, which is 1000.}
\resizebox{0.9\linewidth}{!}{
\begin{tabular}{lccccccc}
\hline
                & WRN    & SS  & ResN-50      & ResN-200     & DeiT-T         & Swin-T           & VMamba-T   \\ 
\hline
epochs          & 200          & 1800         & 270            & 270            & 300            & 300              & 300        \\
WE   & 5            & 5            & 5              & 5              & 5              & 20               & 20        \\
batch size*      & 128$\times$2      & 128$\times$2      & 1024$\times$2       & 1024$\times$2       & 1024$\times$2      & 1024$\times$2         & 1024$\times$2  \\
learning rate   & 0.1          & 0.01         & 0.4            & 0.4            & 1e-3           & 1e-3             & 1e-3      \\
weight decay    & 5e-4         & 1e-3         & 1e-4           & 1e-4           & 0.05           & 0.05             & 0.05      \\ 
\hline
dataset         & CIFAR & CIFAR & ImageNet       & ImageNet       & ImageNet       & ImageNet         & ImageNet  \\
resolution      & 32$\times$32 & 32$\times$32 & 224$\times$224 & 224$\times$224 & 224$\times$224 & 224$\times$224   & 224$\times$224\\ 
\hline
LS & 0            & 0            & 0 (0.1)        & 0 (0.1)        & 0.1            & 0.1              & 0.1   \\
dropout         & 0            & 0            & 0              & 0              & 0              & 0                & 0     \\
droppath        & -            & -            & -              & -              & 0.1            & 0.2              & 0.2   \\
Cutout          & 16           & 16           & -              & -              & -              & -                & -     \\ 
\hline
$\epsilon$      & 2            & 2            & 2              & 2              & $\log c$ (8) & $\log c$ (1) & $\log c$ (1)\\ 
\hline
\end{tabular}
}
\label{hyperparameter-setting}
\end{table*}

\begin{table*}[!h]
\centering
\small
\renewcommand\arraystretch{1.1}
\captionof{table}{The hyperparameters for SRA in the comparing experiments with combined with BA.}
\resizebox{0.7\linewidth}{!}{
\begin{tabular}{lcccc}
\hline
              & WRN & SS & WRN   & SS \\ 
\hline
epochs        & 200       & 200         & 35          & 200         \\
warmup epochs & 5         & 5           & 5           & 5           \\
batch size     & 128$\times$2   & 128$\times$2     & 128$\times$2     & 128$\times$2     \\
learning rate & 0.1       & 0.08        & 0.4         & 0.08        \\
weight decay  & 5e-4      & 1e-3        & 5e-4        & 1e-3        \\ 
\hline
dataset       & CIFAR-10  & CIFAR-10    & CIFAR-100   & CIFAR-100   \\ 
\hline
repeated times & 8        & 8           & 8           & 8           \\
\hline
\end{tabular}
}
\label{BA-hyperparameter-setting}
\end{table*}

\section{Time Cost Evaluation}
\subsection{CIFAR-100}
The total cost is evaluated using WRN-28-10 on CIFAR-100, with batchsize 128$\times$2 for SRA and 128 for the general. We report the sum of search overheads and training cost, with GPU hour as the unit. The basic training time in general is 96 s/epoch on a single RTX 3090, while it is 105 s/epoch for SRA. Therefore, the training cost in general is 96 s/epoch$\times$200 epochs$\approx$5.3 H, while for SRA it is 105 s/epoch$\times$200 epochs$\approx$5.8 H. Methods in the following are listed in the descent of total cost.

\textbf{AA:} The reported search overheads is 5000 H, and the training cost is 5.3 H. Thus, in total it costs 5005.3 H.

\textbf{AdvAA ($\times$8):} As an online search-based method, the search overheads is about 0 H. We omit the time for updating policy parameters. The repeated time for one batch during training is 8, therefore the training cost is estimated as 96 s/epoch$\times$200 epochs$\times$8$\approx$42.7 H. Thus, in total it costs 42.7 H.

\textbf{RA:} No directly reported search overheads. During the search, 5000 samples are left out for evaluation. Five different policy parameter settings are evaluated. Therefore, we estimate the search overheads as 45000/50000$\times$96 s/epoch$\times$200 epochs$\times$5=24.0 H. The training cost is 5.3 H. Thus, in total it costs 29.3 H.

\textbf{MetaA ($\times$4):} As an online search-based method, the search overheads is about 0 H. We omit the time for updating policy parameters. During the search, 1000 samples are left out for evaluation. The reported search epochs is 20, while it takes three times the training time than a standard training scheme \citep{MetaA}. The repeated time for one batch during training is 8, therefore we estimate the training cost as (180 epochs+(20 epochs$\times$3$\times$49000/50000))$\times$96 s/epochs$\times$4$\approx$25.5 H. Thus, in total it costs 25.5 H.

\textbf{BDA:} The reported search overheads is 11.0 H, and the training cost is 5.3 H. Thus, in total it costs 16.3 H.

\textbf{DeepAA:} The reported search overheads is 9.0 H. We ignore the influence of deep augmentation to training cost and estimate it the same as in general, which is 5.3 H. Thus, in total it costs 14.3 H.

\textbf{SLACK:} The reported search overheads is 4.0 H, and the training cost is 5.3 H. Thus, in total it costs 9.3 H.

\textbf{FastAA:} The reported search overheads is 3.5 H, and the training cost is 5.3 H. Thus, in total it costs 8.8 H.

\textbf{AdaAug:} The reported search overheads is 3.3 H, and the training cost is 5.3 H. Thus, in total it costs 8.6 H.

\textbf{SRA ($\times$8):} As a search-free method, the search overheads is 0 H. SRA combined with BA only takes 35 epochs for training, therefore we estimate the training cost as 35 epochs$\times$105 s/epochs$\times$8$\approx$8.2 H. Thus, in total it costs 8.2 H.

\textbf{LA3:} There is no directly reported search overheads. We run the code provided in the paper and evaluate the search overheads as 1.29 H for Stage 1 and 0.02 H for Stage 2. Therefore, the search overheads is about 1.3 H. The training cost is 5.3 H. Thus, in total it costs 6.6 H.

\textbf{FreeA:} The reported search overheads is 1.2 H, and the training cost is 5.3 H. Thus, in total it costs 6.5 H.

\textbf{SRA:} As a search-free method, the search overheads is 0 H. The training cost is 5.8 H. Thus, in total it costs 5.8 H.

\textbf{DDAS:} The reported search overheads is about 0.15 H, and the training cost is 5.3 H. Thus, in total it costs 5.45 H.

\textbf{TA (RA), TA (Wide), and UA:} As search-free methods, the search overheads are 0 H for all of the three. Different search space of TA mainly affects the deformation of augmented images, where the time cost for each epoch is almost the same. Therefore, the training cost for TA (Wide) is also 5.3 H. Thus, in total TA (RA), TA (Wide), and UA cost 5.3 H.

\subsection{ImageNet}
The total costs are evaluated on 4 A100 GPUs using ResNet-50 on ImageNet, with batchsize 1024$\times$2 for SRA and 1024 for the general. The basic training time in general is 540 s/epoch, while it is 591 s/epoch for SRA. Therefore, the training cost in general is 540 s/epoch$\times$270 epochs$\times$4 GPUs=162 H, while for SRA it is 591 s/epoch$\times$270 epochs$\times$4 GPUs=177.3 H. Methods in the following are listed in the descent of total cost.

\textbf{AA:} The reported search overheads is 15000 H, and the training cost is 162 H. Thus, in total it costs 15162 H.

\textbf{RA:} There is no directly reported search overheads. Meanwhile, the models in the original RA paper are only trained for 180 epochs. We estimate that the training strategy follows CIFAR, with 90\% data for training and 10\% left out for evaluation. The total policy parameter settings evaluated by grid search is 6 magnitudes$\times$3 depth=18. We ignore the time cost differences between difference augmentation depth. Therefore, we estimate the search overheads as 90\%$\times$540 s/epoch$\times$180 epochs$\times$4 GPUs$\times$18=1749.6 H. The training cost is 540 s/epoch$\times$180 epochs$\times$4 GPUs$\approx$108 H. Thus, in total it costs 1857.6 H.

\textbf{FastAA:} The model is only trained for 200 epochs. The reported search overheads is 450 H, and the training cost is 540 s/epoch$\times$200 epochs$\times$4 GPUs=120 H. Thus, in total it costs 570 H.

\textbf{DeepAA:} The reported search overheads is 96 H. We ignore the influence of deep augmentation to training cost and estimate the training cost the same as in general, which is 162 H. Thus, in total it costs 258 H.

\textbf{BDA:} The reported search overheads is 28 H, and the training cost is 162 H. Thus, in total it costs 190 H.

\textbf{SRA:} As a search-free method, the search overheads is 0 H. The training cost is 177.3 H. Thus, in total it costs 177.3 H.

\textbf{DDAS:} The reported search overheads is about 1.3 H, and the training cost is 162 H. Thus, in total it costs 163.3 H.

\textbf{TA (Wide), and UA:} As search-free methods, the search overheads are 0 H for both the two, while the training cost is also 162 H. Thus, in total TA (Wide) and UA cost 162 H.

\end{document}